\documentclass{llncs}

\usepackage{subcaption}
\usepackage{comment}
\usepackage{amsmath}
\usepackage{amssymb}
\usepackage{amsfonts}
\usepackage{graphicx}
\usepackage{color}
\usepackage{multicol}
\usepackage{multirow}
\usepackage[ruled,noline]{algorithm2e}
\usepackage{comment}
\usepackage{booktabs}
\usepackage{subcaption}
\usepackage{lipsum}
\newcommand{\bumpup}{\vspace*{-2.0ex}}

\usepackage{booktabs}

\usepackage{graphicx}
\begin{document}
\title{Local block-wise self attention for normal organ  segmentation} 
%
%
\author{Jiang Jue\inst{1}
	\and  Sharif Elguindi \inst{1} 
	\and Um Hyemin\inst{1}
	\and Sean Berry\inst{2}
	\and Veeraraghavan Harini\inst{1} }

\institute{Medical Physics, Memorial Sloan Kettering Cancer Center \and Radiation Oncology, Memorial Sloan Kettering Cancer Center}

%
%

\maketitle              
\begin{abstract}
We developed a new and computationally simple local block-wise self attention based normal structures segmentation approach applied to head and neck computed tomography (CT) images. Our method uses the insight that normal organs exhibit regularity in their spatial location and inter-relation within images, which can be leveraged to simplify the computations required to aggregate feature information. We accomplish this by using local self attention blocks that pass information between each other to derive the attention map. We show that adding additional attention layers increases the contextual field and captures focused attention from relevant structures. We developed our approach using U-net and compared it against multiple state-of-the-art self attention methods. All models were trained on 48 internal headneck CT scans and tested on 48 CT scans from the external public domain database of  computational anatomy dataset.  Our  method achieved the highest Dice similarity coefficient segmentation accuracy of 0.85$\pm$0.04, 0.86$\pm$0.04 for left and right parotid glands, 0.79$\pm$0.07 and 0.77$\pm$0.05 for left and right submandibular glands, 0.93$\pm$0.01 for mandible and 0.88$\pm$0.02 for the brain stem with the lowest increase of 66.7\% computing time per image and 0.15\% increase in model parameters compared with standard U-net. The best state-of-the-art method called point-wise spatial attention, achieved \textcolor{black}{comparable accuracy but with 516.7\% increase in computing time and 8.14\% increase in parameters compared with standard U-net.} Finally, we performed ablation tests and studied the impact of attention block size, overlap of the attention blocks, additional attention layers, and attention block placement on segmentation performance.


\end{abstract}

\section{Introduction}
\bumpup

Notes to self: Self-attention is a mechanism to identify auxiliary objects to help improve detection accuracy of target structures. These are objects that generated from the saliency map of the region surrounding the target structures, and correspond to those that are highly likely to have an interaction with the target. 
Auxiliary objects appear with the target. Our goal here is to identify the auxiliary candidates that increase the detection performance of the target structures.

 Computed tomography (CT) is the standard imaging modality used in radiation treatment planning. However, low soft-tissue contrast in CT~\cite{whitfield2013automated}, restricts the segmentation accuracy that can be achieved for various soft-tissue organ-at-risk structures. Advanced methods developed for medical images typically combine features computed from deeper layers~\cite{dou2018local,oktay2018} to focus and improve segmentation.
 \begin{figure}
\centering
\includegraphics[width=0.6\textwidth]{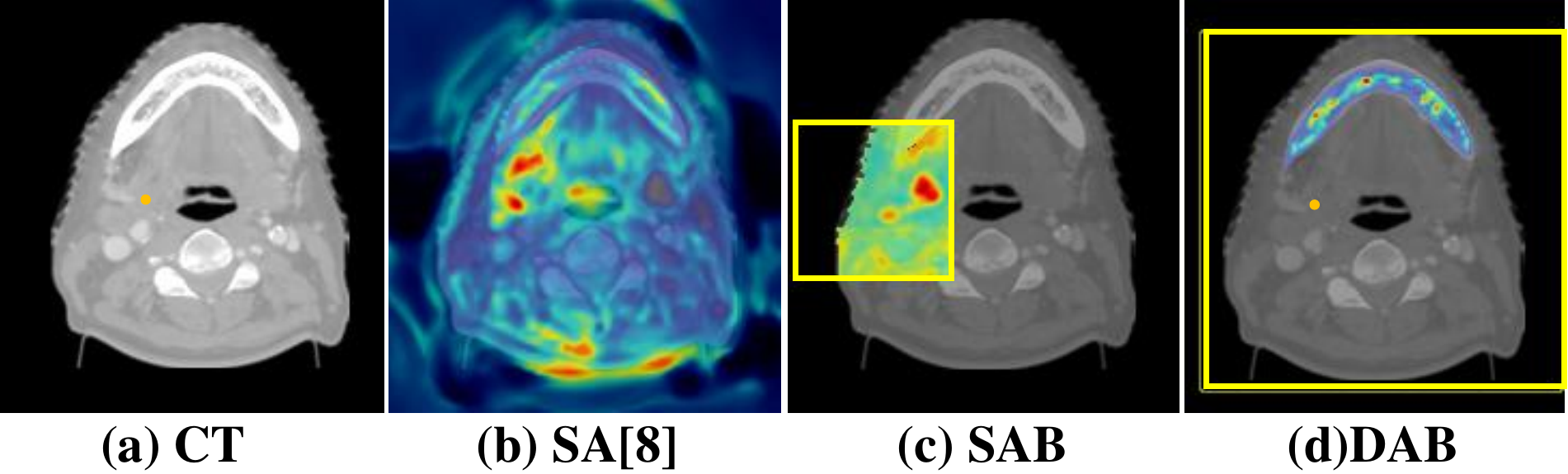}
\caption{\small{Comparison between (B) non-local (SA) and block-wise self attention using single attention block (C) and dual attention block (D) layer. Yellow rectangle indicates the effective contextual field; yellow dot represents a submandibular glands pixel.}} \label{fig:motivation}
\end{figure}
Recent developments in self attention networks~\cite{parmar2018image} enables aggregation of long-range contextual information that has been shown to produce more accurate segmentations in real-world~\cite{huang2018ccnet,fu2018dual} images.  Self attention aggregates features from all pixels within a context such as the whole image features  in order to build support for a particular voxel. Such feature aggregation requires intensive computations to model the long-range dependencies~\cite{vaswani2017attention,wang2018non}. Therefore, we developed a new approach that employs local computations using 2D self attention blocks.\\
Our approach uses the insight that normal organs exhibit regularity in their spatial location and relation with respect to one another. Our approach exploits this regularity to simplify feature aggregation by processing information from local self attention blocks to model the pixel context. These blocks pass information between each other as the image (or its feature representation) is scanned in raster-scan order to then derive the attention flow. We also developed a stacked attention formulation by adding additional attention layers that improves inference by both increasing the contextual view and by capturing multiple information from different parts of the image~\cite{wang2018}.
\\
Prior works attempted to reduce the computational burden for self attention by modeling relations between objects in an image~\cite{hu2018relation}, by successive pooling~\cite{yuan2018ocnet}, and by aggregating  information spatially and from features channels~\cite{fu2018dual}. The approach  in~\cite{huang2018ccnet} reduced computations by considering only the pixels lying in the horizontal and vertical directions of each pixel. However, this approach also ignores relations between pixels that occur in diagonal orientations. 
\\
Our approach improves on the segmentation and computational performance of prior methods as shown by our results. Figure~\ref{fig:motivation} shows an example case with self attention map generated for a pixel (indicated by a yellow dot) randomly placed within the submandibular glands (Figure~\ref{fig:motivation}A) using the non-local network~\cite{wang2018non}, herefrom referred as SA (Figure~\ref{fig:motivation}B), and our method using \underline{s}ingle layer of \underline{a}ttention \underline{b}lock (SAB) (Figure~\ref{fig:motivation}C) and two layers or \underline{d}ual \underline{a}ttention \underline{b}locks (DAB) (Figure~\ref{fig:motivation}D). As seen, the attention maps derived from our approach tends to focus within local structures and also captures information from relevant structures. For reproducible research, we will make the code available upon acceptance. 

\begin{figure}
\centering
\includegraphics[width=1\textwidth,scale=0.3]{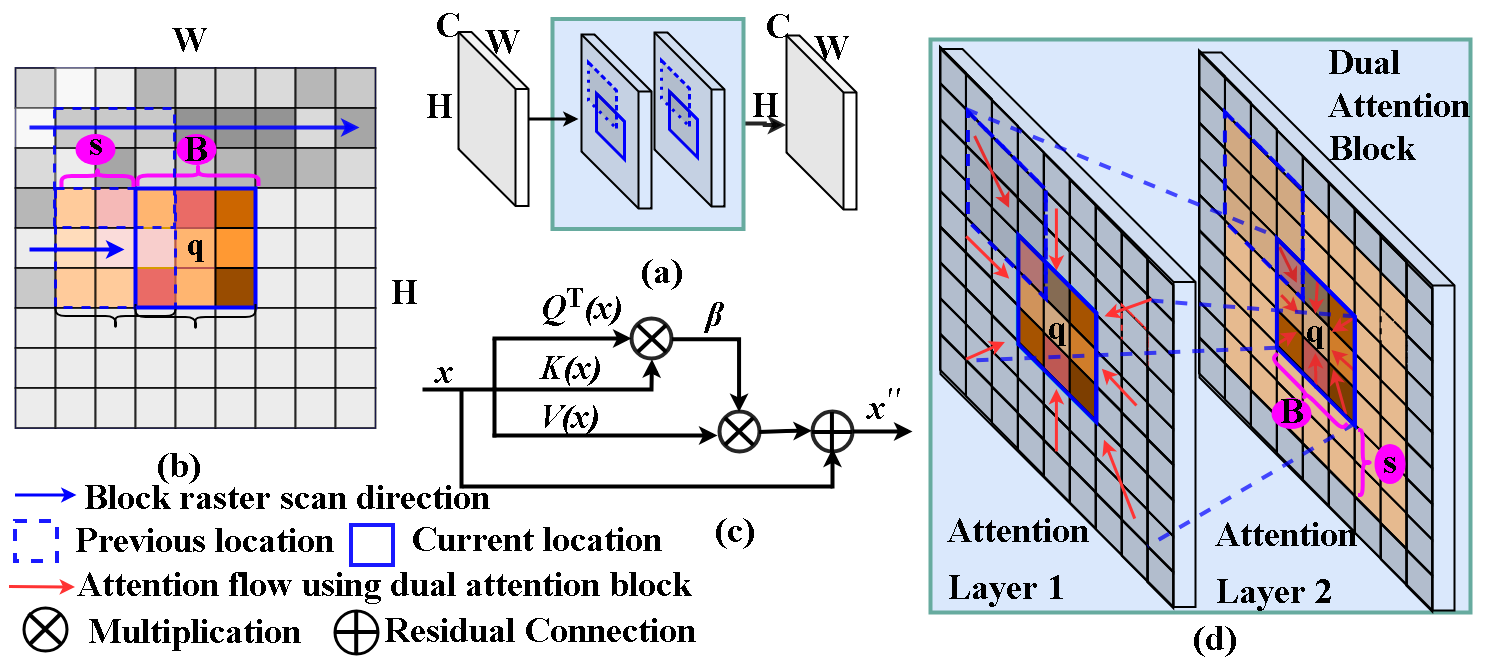}
\caption{\small {Illustration of block-wise self attention connected with U-net shown in (a), scanning using block-wise self attention (b), schematic of self attention in (c), and stacked self attention in (d) helps to increase contextual field. Attention blocks (blue rectangle) are computed in  top-down raster scan order.}} \label{fig:block}
\end{figure}
\bumpup
\section{Method}
\bumpup
Given an image $I$, the goal of the proposed method is to produce a segmentation $S$ corresponding to one or more structures in a computationally fast manner and with little memory requirement. We achieve this by using local attention blocks. An attention block is a region within which local self attention is computed. Addition of multiple attention layers enables attention to flow and increases the contextual field, thereby, modelling long-range contextual information (Figure~\ref{fig:block}).  
\bumpup
\subsection{Self Attention}
\bumpup
In standard self attention~\cite{vaswani2017attention}, given an image represented by feature $x$ $\in$ $\mathbb{R}^{C \times W \times H}$, where $C$, $W$, $H$ denote the size of feature channel, width and height, three feature embeddings of size $C^{'} \times W \times H$, where $C^{'} \leq C$, corresponding to query $Q(x)$, key $K(x)$, and value $V(x)$ are computed by projecting the feature through $1 \times 1$ convolutional filters (Figure~\ref{fig:block} C). The attention map is calculated by taking a matrix product of the query and the key features as:
\begin{equation}
\begin{split}
\beta_{j,i}=\frac{exp(s_{i,j})}{\sum_{i=1}^{N}exp(s_{ij})},\textrm{where} \quad    
 s_{ij}=Q(x_{i})^{T}K(x_{j}),
\end{split}
\end{equation}
where $\beta_{j,i}$ is the result of softmax computation and measures the impact of feature at position $i$ on the feature at position $j$, whereby, similar features give rise to higher attention values. Thus attention for each feature roughly corresponds to its similarity to all other features in the context. The attention aggregated feature representation is computed by multiplying with value $V$ as:
\begin{equation}
\begin{split}
x^{''}_{j}=\sum_{i=1}^{N}\beta_{j,i}V(x_{i}) + x_{j}.
\end{split}
\end{equation}
Computing a global self attention map $\beta$ requires HW$\times$HW computations to cover an image feature of size $C \times H \times W$, which in turn leads to time and space complexity of ($O(H^4)$), where ($H \geq  W$). This can quickly become prohibitive even for standard medical image volumes. Our approach simplifies these computations as described in the following subsection. 
\bumpup
\subsection{Local block-wise self attention}
\bumpup
Our approach is similar to that in~\cite{parmar2018image} for image generation. However, instead of employing an encoder for each pixel, which ignores the correlation between spatially adjacent pixels, we used 2D convolutions to jointly encode multiple pixels. Our approach reduces the computations compared with global self-attention methods by focusing feature aggregation to within fixed size attention blocks. When using non-overlapping attention blocks, an image of size $H \times W $ and represented by features $x \in \mathbb{R}^{C \times W \times H}$ is divided into ($\frac{W \times H}{B \times B}$) blocks by scanning in a raster-scan order, where $B \times B$ is the size of the attention block. Non-overlapping attention blocks may result in block-like artifacts on the attention map. Therefore, we use overlapping attention blocks with stride $s, s < B$ (Figure~\ref{fig:block}(b)). Overlapping strides also facilitates information flow when stacking attention layers (red arrow in Figure 2d) as described below. The number of computations required to calculate attention of a single block is  $(B)^{2}$$\times$$(B)^{2}$, which corresponds to  $(B)^{2}$$\times$$(B)^{2}$$\times$$(\frac{H-(B-s)}{s})$$\times$$(\frac{W-(B-s)}{s})$\footnote{\textcolor{black}{We assume an image feature is divisible otherwise padding is required}} for the whole image.
\\
 Using attention blocks restricts the contextual field (within which feature is aggregated) to a $B \times B$ region for one attention layer. Therefore, we increase the contextual field to $(B+2s) \times (B+2s)$ by adding a second attention layer (layer 2 in Figure 2d). We call single layer attention as single attention block (SAB) and two layer attention as dual attention block (DAB). Each addition of attention layer doubles the computation cost for each block by $n \times B^{4}$, where $n$ is the number of attention layers and $B \ll min(H,W)$. 

\bumpup
\subsection{Implementation and Network Structure}
\bumpup
All networks were implemented using the Pytorch library ~\cite{paszke2017automatic} and  trained on Nvidia GTX 1080Ti with 12GB memory. The ADAM algorithm ~\cite{kingma2014adam} with an initial learning rate of 2e-4 was used during training. We implemented the attention method using U-net\cite{ronneberger2015u}. We modified U-net to include batch normalization after each convolution layer to speed up convergence. 

\bumpup
\section{Datasets and evaluation metrics}
\bumpup
A total of 96 head and neck CT datasets were analyzed. All networks were trained using 48 patients from internal archive and tested on 48 patients from the external public domain database for computational anatomy (PDDCA)\cite{raudaschl2017evaluation}. Training was performed by cropping the images to the head area and using image patches resized to 256 $\times$ 256, resulting in a total of 8000 training images. Models were constructed to segment multiple organs present in both datasets that included parotid glands, submandibular glands, brain stem, and manidble. Segmentation accuracy was evaluated by comparing with expert delineations using the Dice similarity coefficient (DSC). 

\bumpup
\begin{figure}
\centering
\includegraphics[width=0.83\textwidth]{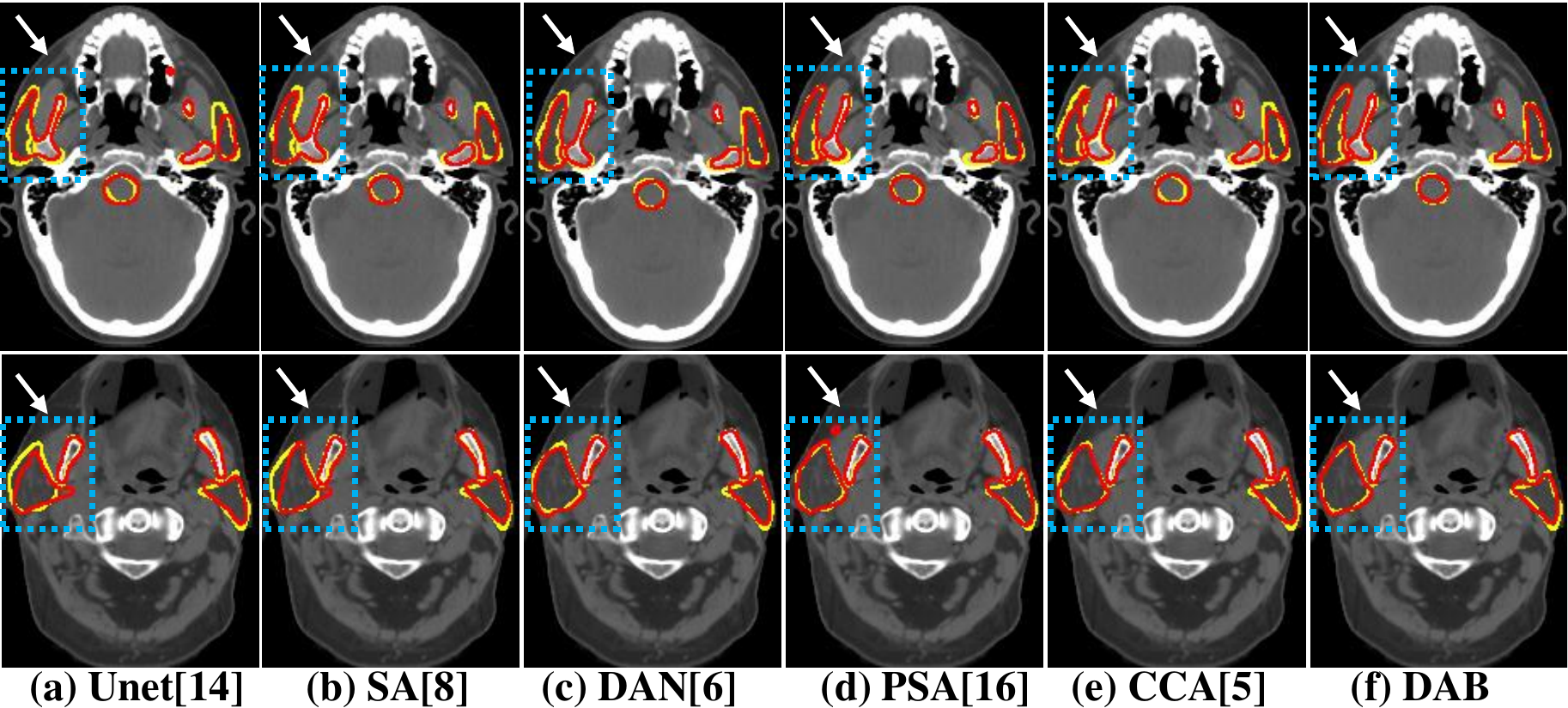}
\caption{\small {Comparison of segmentations produced by our and competitive methods. Expert contours are shown in yellow and algorithm segmentations are in red. The blue box indicates those parts with segmentation discrepancy between algorithm and expert.}} \label{fig:seg_overlay}
\end{figure}

\section{Experiments}
\bumpup
We compared our method with the following deep learning self attention modules proposed in: (i) non-local neural network  (SA)~\cite{wang2018non}, (ii) Dual attention net (DAN)~\cite{fu2018dual}, (iii) Point-wise spatial attention (PSA)~\cite{zhao2018psanet} and (iv) criss-cross attention (CCA)~\cite{huang2018ccnet}. Default settings included a block size of $36\times36$, a scanning stride of 24 with dual layer attention implemented on the penultimate layer\footnote{\textcolor{black}{a set of computations} consisting of CONV, BN, Relu is treated as a layer.} of U-net with feature size 64$\times$128$\times$128 (C$\times$H$\times$W). Details of attention block placement are listed in supplementary document. For equitable comparison, other modules or methods were also implemented on the penultimate layer. We set self attention feature channel embedding  $C^{'}$=C$/$2=32. Ablation tests were conducted to study the influence of (a) single vs. multiple attention layers, (b) placement of attention blocks (penultimate vs. last layer), (c) attention block sizes (B=24, 36, 48), and (d) overlapping \textcolor{black}{($s$=$B$$\times$$\frac{2}{3}$)} vs. non-overlapping \textcolor{black}{($s$=$B$)} attention blocks.

\section{Results}
\bumpup
\subsection{Segmentation accuracy}
\bumpup
Table~\ref{tab:segmentationUnet} shows a comparison of the segmentation accuracy achieved by the various methods using mean and standard deviation, computational complexity, number of model parameters with \% increase in number of parameters compared with standard U-net ($\Delta m$), the computing time measured as an average during training, and the \% increase in computations over U-net ($\Delta t$). Our method (DAB) produced the most accurate segmentation for all the analyzed organs. It required fewer computations and parameters compared with all except the criss-cross (CCA) method.  DAB was also the fastest to compute among the self attention methods. The multiplication by 2 for complexity in DAB and CCA are due to the addition of a second attention layer. Figure~\ref{fig:seg_overlay} shows two representative examples with the algorithm and expert delineation. The arrows indicate problematic segmentations. As seen, whereas U-net and SA resulted in over-segmentation of the parotid gland (top row) and multiple methods resulted in under-segmentation of the right parotid gland (bottom row), DAB closely approximated the expert delineation.

\bumpup
\begin{table} 
	\centering{\caption{\small {Comparison of segmentation performance between proposed and competitive methods. Analyzed structures: left parotid-LP, right parotid gland -RP, left submandibular-LS, right submandibular gland-RS, mandible-M, and brain stem-BS.}} 
		\label{tab:segmentationUnet} 
		\centering
		\footnotesize
		
		\begin{tabular}{|c|c|c|c|c|c|c|c|c|c|} 

			\hline 
			{  Method  }  & {  LP  }  & {  RP }& {  LS  }& {  RS  }& {  M }&{  BS  }&{Complexity}&{Param(M)/$\Delta m$}&{secs/$\Delta t$}\\
			\hline 
			\multirow{2}{*}{U-net\cite{ronneberger2015u}}  & {  0.81  }  & {  0.83 }& {  0.77  }& {  0.73  }& {  0.91 }&{  0.87  }&{N/A}&{13.39/-}&{0.06/-}\\	
			{}&{0.07}& {  0.06 }& {  0.07  }& {  0.09  }& {  0.02 }&{  0.02  }&{N/A}&{}&{}\\	
			\hline 
			\multirow{2}{*}{\textrm{+}SA\cite{wang2018non}}  & {  0.83  }  & {  0.84 }& {  0.79  }& {  0.76  }& {  0.92 }&{  0.87  }&{128$\times$128$\times$}&{13.43/0.30\%}&{0.16/166.7\%}\\	
				{}&{0.05}& {  0.06 }& {  0.07  }& {  0.07  }& {  0.03 }&{  0.02  }&{128$\times$128}&{}&{}\\
			\hline 
			\multirow{2}{*}{+DAN\cite{fu2018dual}}  & {  0.83  }  & {  0.84 }& {  0.79  }& {  0.77  }& {  0.93 }&{  0.87  }&{256$\times$128$\times$}&{13.46/0.52\%}&{0.19/216.7\%}\\	
				{}&{0.06}& {  0.05 }& {  0.06  }& {  0.07  }& {  0.02 }&{  0.03  }&{128$\times$128  }&{}&{}\\
			\hline 
			\multirow{2}{*}{+PSA\cite{zhao2018psanet}}  & {  0.84  }  & {  0.84 }& {  0.79  }& {  0.77  }& {  0.93 }&{  0.87  }&{128$\times$128$\times$ }&{14.48/8.14\%}&{0.37/516.7\%}\\	
				{}&{0.05}& {  0.04 }& {  0.07  }& {  0.06  }& {  0.01 }&{  0.02  }&{255$\times$255 }&{}&{}\\
			\hline 
			\multirow{2}{*}{+CCA\cite{huang2018ccnet}}  & {  0.83  }  & {  0.84 }& {  0.78  }& {  0.76  }& {  0.93 }&{  0.87  }&{\textbf{128$\times$128$\times$} }&{13.41/0.15\%}&{0.16/166.7\%}\\	
				{}&{0.07}& {  0.06 }& {  0.08  }& {  0.08  }& {  0.02 }&{  0.02  }&{\textbf{255 $\times$ 2} }&{}&{}\\
			\hline 
			\multirow{2}{*}{+DAB }  & {  0.85  }  & {  0.86 }& {  0.79  }& {  0.77  }& {  0.93 }&{  0.88 }&{$36^{2}$$\times$ $36^{2}$ $\times$}&{13.41/0.15\%}&{0.10/66.7\%}\\	
				{}&{0.04}& {  0.04 }& {  0.07  }& {  0.05  }& {  0.01 }&{  0.02  }&{\textcolor{black}{5$\times 5  \times$2}}&{}&{}\\
			\hline 

	\end{tabular}} 
\end{table}
\subsection{Ablation experiments}
\textbf{Attention layers:} As shown in Table \ref{tbl:iterNum_P} and \ref{tbl:iterNum_L}, addition of attention block layers in the penultimate layer improves the segmentation performance with little additional computational time. Placement of attention in the last layer slightly increased accuracy. Besides CCA, its infeasible to add single attention in the last layer using other methods due to memory limitations. \\
\textbf{Attention block size:} There is only a minimal difference in the segmentation accuracy by increasing the block sizes (see Table \ref{tbl:over_P},\ref{tbl:over_L}). \\
\textbf{\textcolor{black}{Overlapping vs. non-overlapping attention blocks:}} The difference of overlapping and non-overlapping become more obvious, namely with higher accuracy achieved with a smaller block size for overlapping blocks, especially when attention blocks were placed in the last layer. 

\bumpup
\subsection{Qualitative results using attention map}
\begin{figure}
\centering
\includegraphics[width=0.8\textwidth]{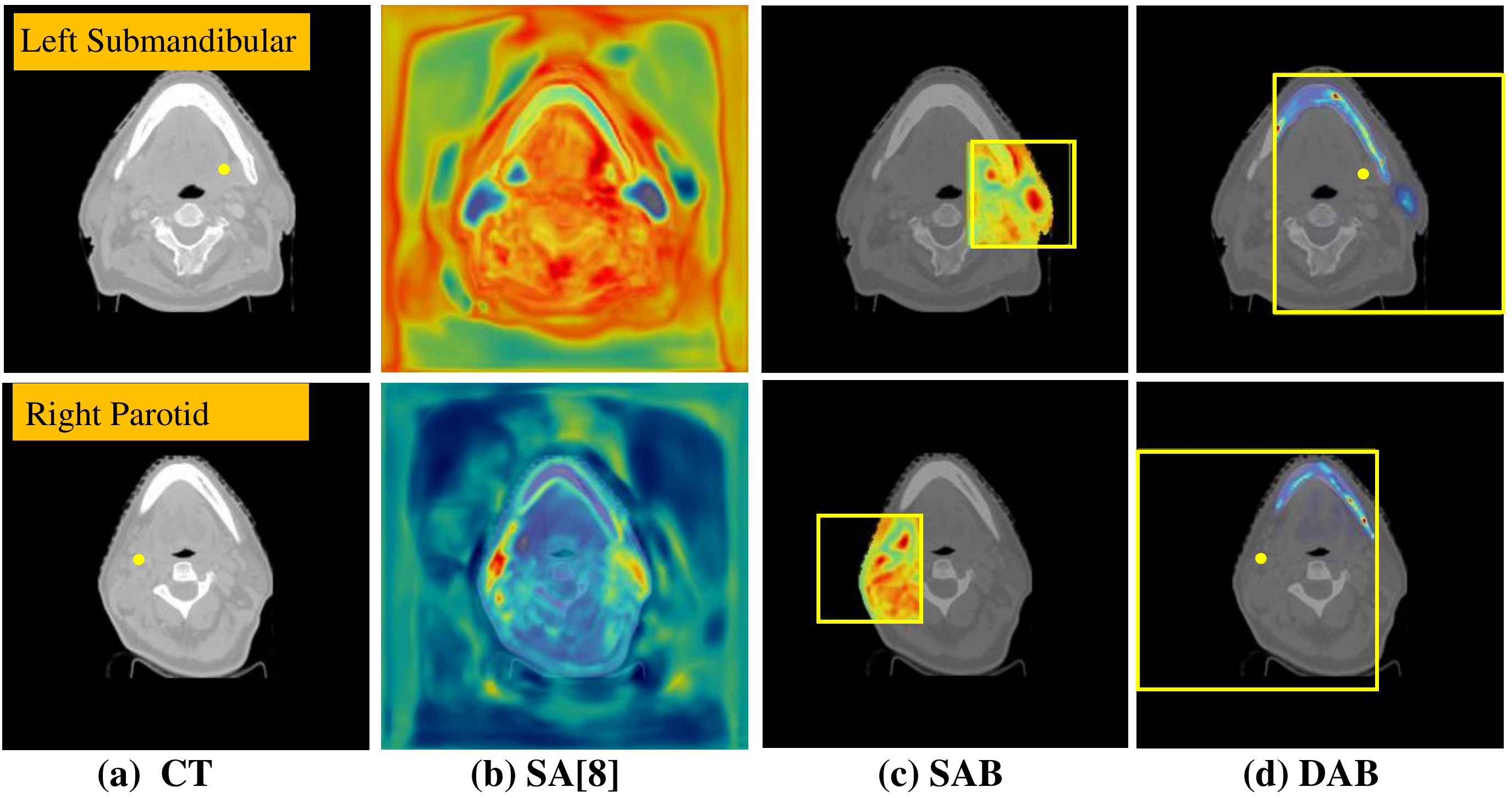}
\caption{\small Attention maps for representative cases for for interest point (shown in yellow) using (b) self attention (SA)\cite{wang2018non}, (b) single attention block (SAB) and (c) dual attention block (DAB) layer. Yellow rectangle indicates the effective contextual field.} \label{fig:attention map}
\end{figure}

\bumpup
Figure~\ref{fig:attention map} shows attention maps for representative examples computed by using SAB and DAB placed on the penultimate layer. As shown, changing the number of attention  layers from one (SAB) to two (DAB) changes and increases the context of the structures involved, from local in SAB to adjacent and relevant structures in DAB. For comparison, the attention map computed using the SA method~\cite{wang2018non} is also shown. As shown, the SA~\cite{wang2018non} map includes all portions of the image, which does not lead to improved performance (Table~\ref{tab:segmentationUnet}). 

\bumpup
\section{Discussions}
\bumpup
We developed a computationally efficient approach that achieved similar to better performance than state-of-the-art self attention methods for normal organ segmentation in head and neck CT scans. While computing a global attention such as in prior methods ~\cite{wang2018non,zhao2018psanet} clearly have the benefit of incorporating information for structures that have large variability in their location such as in scene parsing, such methods do not necessarily lead to improved performance for normal organ segmentation compared to block-wise attention. 

\bumpup
\begin{table}[tbp]
\centering
\begin{subtable}[t]{0.48\linewidth}
\centering
\vspace{0pt}
		\scriptsize
		
\begin{tabular}{|c|c|c|c|c|c|c|c|} 

			\hline 
			{  N   }  & {  LP  }  & {  RP }& {  LS  }& {  RS  }& {  M }&{  BS  }&{secs}\\
			\hline 
			\multirow{2}{*}{1}  &  {  0.83  }  & {  0.84 }& {  0.78  }& {  0.76  }& {  0.92 }&{ 0.87  }&{0.09}\\
			
            {  }  & {  0.05  }  & {  0.04 }& {  0.06  }& {  0.05  }& {  0.01 }&{ 0.02  }&{}\\			
			\hline 
			\multirow{2}{*}{2}  &  {  \textbf{0.85}  }  & {  \textbf{0.86} }& {  \textbf{0.79}  }& {  0.77  }& {  \textbf{0.93} }&{ \textbf{0.88}  }&{0.10}\\
			
            {  }  & {  0.04  }  & {  0.04 }& {  0.07  }& {  0.05  }& {  0.01 }&{ 0.02  }&{}\\	
            \hline
		\multirow{2}{*}{3}  &  {  0.84  }  & {  0.85 }& {  0.79  }& {  \textbf{0.78}  }& {  0.93 }&{ 0.88  }&{0.15}\\
			
            {  }  & {  0.05  }  & {  0.05 }& {  0.08  }& {  0.05  }& {  0.01 }&{ 0.02  }&{}\\	
            \hline
            
	\end{tabular} 
\caption{\small{Number (N) of attention blocks placed on the \textbf{penultimate layer}.}}\label{tbl:iterNum_P}

\end{subtable}\hfill
\begin{subtable}[t]{0.48\linewidth}
\centering
\vspace{0pt}
		\scriptsize
		
\begin{tabular}{|c|c|c|c|c|c|c|c|} 

			\hline 
			{  N   }  & {  LP  }  & {  RP }& {  LS  }& {  RS  }& {  M }&{  BS  }&{secs}\\
			\hline 
			\multirow{2}{*}{1}  &  {  0.85  }  & {  0.85 }& {  0.78  }& {  0.75  }& {  0.92 }&{ 0.88  }&{0.24}\\
			
            {  }  & {  0.05  }  & {  0.04 }& {  0.08  }& {  0.09  }& {  0.02 }&{ 0.02  }&{}\\			
			\hline 
			\multirow{2}{*}{2}  &  {\textbf{0.85}}  & {\textbf{0.86} }& { \textbf{0.79} }& {  \textbf{0.79} }& { \textbf{ 0.93} }&{ \textbf{0.88}}&{0.44}\\
			
            {  }  & {  0.04  }  & {  0.04 }& {  0.07  }& {  0.06  }& {  0.02 }&{ 0.02  }&{}\\	
            \hline
		\multirow{2}{*}{3}  &  { 0.85  }  & {  0.86 }& {  0.79  }& {  0.79  }& {  0.93 }&{ 0.88  }&{0.65}\\
			
            {  }  & {  0.04  }  & {  0.04 }& {  0.06  }& {  0.05  }& {  0.02 }&{ 0.02  }&{}\\	
            \hline
	\end{tabular} 
\caption{\small{Number (N) of attention blocks on \textbf{last layer}.}}\label{tbl:iterNum_L}

\end{subtable}\hfill

\begin{subtable}[t]{0.48\linewidth}
\centering
\vspace{0pt}
		\scriptsize
		
\begin{tabular}{|c|c|c|c|c|c|c|c|} 

			\hline 
			\multicolumn{8}{|c|}{Overlap} \\
			\hline
			{B}  & {  LP  }  & {  RP }& {  LS  }& {  RS  }& {  M }&{  BS  }&{secs}\\
			\hline 
			
			\multirow{2}{*}{24}  &  {  0.84  }  & {  0.85 }& {  0.79  }& {  0.77  }& {  0.93 }&{ 0.88  }&{0.23}\\
			
            {  }  & {  0.06  }  & {  0.05 }& {  0.07  }& {  0.06  }& {  0.02 }&{ 0.02  }&{}\\			
			\hline 			
					\multirow{2}{*}{36}  &  {  \textbf{0.85}  }  & { \textbf{0.86} }& {  \textbf{0.79}  }& {  \textbf{0.77}  }& {  \textbf{0.93} }&{ \textbf{0.88}  }&{0.10}\\
			
            {  }  & {  0.04  }  & {  0.04 }& {  0.07  }& {  0.05  }& {  0.01 }&{ 0.02  }&{}\\	
			\hline
			\multirow{2}{*}{48}  &  {  0.85  }  & {  0.85 }& {  0.79  }& {  0.77  }& {  0.93 }&{ 0.88  }&{0.13}\\
			
            {  }  & {  0.05  }  & {  0.05 }& {  0.06  }& {  0.05  }& {  0.01 }&{ 0.01  }&{}\\			
			\hline 
			\hline 
			\multicolumn{8}{|c|}{Non-Overlap} \\
			\hline
			{B}  & {  LP  }  & {  RP }& {  LS  }& {  RS  }& {  M }&{  BS  }&{secs}\\
			\hline 
			
			\multirow{2}{*}{24}  &  {  0.84  }  & {  0.84 }& {  0.78  }& {  0.77  }& {  0.93 }&{ 0.88  }&{0.16}\\
			
            {  }  & {  0.05  }  & {  0.05 }& {  0.07  }& {  0.05  }& {  0.01 }&{ 0.02  }&{}\\			
			\hline 			
					\multirow{2}{*}{36}  &  {  \textbf{0.84}  }  & {  \textbf{0.85} }& {  \textbf{0.79}  }& {  \textbf{0.77}  }& {  \textbf{0.93} }&{ \textbf{0.88}  }&{0.09}\\
			
            {  }  & {  0.05  }  & {  0.04 }& {  0.07  }& {  0.06  }& {  0.01 }&{ 0.02  }&{}\\	
			\hline
			\multirow{2}{*}{48}  &  {  0.84  }  & {  0.85 }& {  0.79  }& {  0.77  }& {  0.93 }&{ 0.87  }&{0.09}\\
			
            {  }  & {  0.06  }  & {  0.06 }& {  0.06  }& {  0.06  }& {  0.01 }&{ 0.03  }&{}\\			
			\hline

			
	\end{tabular} 
\caption{\small{\textbf{Overlapping} blocks with varying sizes (B) in \textbf{penultimate layer}.}}\label{tbl:over_P}

\end{subtable}\hfill
\begin{subtable}[t]{0.48\linewidth}
\centering
\vspace{0pt}
		\scriptsize
\begin{tabular}{|c|c|c|c|c|c|c|c|} 
            \hline
            \multicolumn{8}{|c|}{Overlap} \\
			\hline 
			{B}  & {  LP  }  & {  RP }& {  LS  }& {  RS  }& {  M }&{  BS  }&{secs}\\
			\hline 
			
			\multirow{2}{*}{24}  &  {  0.85  }  & {  0.86 }& {  0.79  }& {  0.78  }& {  0.93 }&{ 0.88  }&{0.88}\\
			
            {  }  & {  0.04  }  & {  0.04 }& {  0.06  }& {  0.06  }& {  0.01 }&{ 0.02  }&{}\\			
			\hline 			
			\multirow{2}{*}{36}  &  {  \textbf{0.85}  }  & {  \textbf{0.86} }& {  \textbf{0.79}  }& { \textbf{0.79}  }& {  \textbf{0.93} }&{ \textbf{0.88}  }&{0.44}\\
			
            {  }  & {  0.04  }  & {  0.04 }& {  0.07  }& {  0.06  }& {  0.02 }&{ 0.02  }&{}\\			
			\hline 	
			\multirow{2}{*}{48}  &  { -  }  & {  - }& {  -  }& { -  }& {  - }&{ -  }&{-}\\
			
            {  }  & {  -  }  & {  - }& {  - }& {  -  }& { - }&{-  }&{-}\\			
			\hline 
			\hline 
			\multicolumn{8}{|c|}{Non-Overlap} \\
			\hline
	{B}  & {  LP  }  & {  RP }& {  LS  }& {  RS  }& {  M }&{  BS  }&{secs}\\
			\hline 			
			\multirow{2}{*}{24}  &  {  0.85  }  & {  0.85 }& {  0.77  }& {  0.76  }& {  0.92 }&{ 0.88  }&{0.42}\\
			
            {  }  & {  0.05  }  & {  0.04 }& {  0.09  }& {  0.07  }& {  0.02 }&{ 0.02  }&{}\\			
			\hline 			
			\multirow{2}{*}{36}  &  {  \textbf{0.84}  }  & {  \textbf{0.85} }& {  \textbf{0.79}  }& {  \textbf{0.78}  }& {  \textbf{0.93} }&{ \textbf{0.88}  }&{0.23}\\
			
            {  }  & {  0.05  }  & {  0.05 }& {  0.06  }& {  0.06  }& {  0.01 }&{ 0.02  }&{}\\			
			\hline 	
			\multirow{2}{*}{48}  &  { -  }  & {  - }& {  -  }& { -  }& {  - }&{ -  }&{}\\
			
            {  }  & {  -  }  & {  - }& {  - }& {  -  }& { - }&{-  }&{}\\			
			\hline 				
			
	\end{tabular} 
\caption{\small{\textbf{Overlapping} blocks with varying sizes (B) in \textbf{last layer}.}}\label{tbl:over_L}

\end{subtable}\hfill

\caption{\small{Ablation tests to assess segmentation accuracy and computation time. } }\label{tbl:ablation}
\end{table}

\bumpup
\section{Conclusion}
\bumpup
We developed a novel block-wise self attention approach for segmenting normal organ structures from head and neck CT scans. Our results show that our method achieves computationally more efficient and better segmentation than multiple state-of-the art self attention methods. 

{\tiny
	\bibliographystyle{splncs}
}
\bibliography{refs}
%
%
%
%





\end{document}